\documentclass[11pt]{article}

\PassOptionsToPackage{table,xcdraw}{xcolor}

\usepackage[final]{acl}

\usepackage{times}
\usepackage{latexsym}

\usepackage[T1]{fontenc}

\usepackage[utf8]{inputenc}

\usepackage{microtype}

\usepackage{inconsolata}

\usepackage{graphicx}
\usepackage{amsmath}
\usepackage{amssymb}
\usepackage{booktabs}

\usepackage{tabularx}
\usepackage{algorithm}
\usepackage{algpseudocode}

\usepackage{tabularx}
\usepackage{algorithm}
\usepackage{algpseudocode}

\definecolor{open_models_below_4B}{RGB}{185, 235, 255}
\definecolor{open_models_below_12B}{RGB}{255, 219, 187}
\definecolor{open_models_over_12B}{RGB}{190, 255, 190}
\definecolor{closed_models}{RGB}{240, 240, 240}

\usepackage{listings}
\usepackage{subcaption}

\usepackage{markdown}

\colorlet{punct}{red!60!black}
\definecolor{background}{HTML}{EEEEEE}
\definecolor{delim}{RGB}{20,105,176}
\colorlet{numb}{magenta!60!black}
\lstdefinelanguage{json}{
    showstringspaces=false,
    breaklines=true,
    frame=single,
    backgroundcolor=\color{background},
    literate=
     *{0}{{{\color{numb}0}}}{1}
      {1}{{{\color{numb}1}}}{1}
      {2}{{{\color{numb}2}}}{1}
      {3}{{{\color{numb}3}}}{1}
      {4}{{{\color{numb}4}}}{1}
      {5}{{{\color{numb}5}}}{1}
      {6}{{{\color{numb}6}}}{1}
      {7}{{{\color{numb}7}}}{1}
      {8}{{{\color{numb}8}}}{1}
      {9}{{{\color{numb}9}}}{1}
      {:}{{{\color{punct}{:}}}}{1}
      {,}{{{\color{punct}{,}}}}{1}
      {\{}{{{\color{delim}{\{}}}}{1}
      {\}}{{{\color{delim}{\}}}}}{1}
      {[}{{{\color{delim}{[}}}}{1}
      {]}{{{\color{delim}{]}}}}{1},
}

\newcommand{\ourmodel}{{\textsc{StarFlow}}}

\title{\textsc{StarFlow}: Generating Structured Workflow Outputs From Sketch Images}

\author{%
Patrice Bechard$^{1}$ \quad
Chao Wang$^{1}$ \quad
Amirhossein Abaskohi$^{1,2}$ \\
\textbf{Juan Rodriguez}$^{1,3,4}$ \quad
\textbf{Christopher Pal}$^{1,3,5,6}$ \quad
\textbf{David Vazquez}$^{1}$ \\
\textbf{Spandana Gella}$^{1}$\thanks{Equal supervision.} \quad
\textbf{Sai Rajeswar}$^{1,3}$\footnotemark[1] \quad
\textbf{Perouz Taslakian}$^{1}$\footnotemark[1] \\
$^1$ServiceNow \quad  $^2$University of British Columbia \quad $^3$Mila \\ $^4$École de Technologie Supérieure \quad $^5$CIFAR AI Chair \quad $^6$Polytechnique Montréal
}

\begin{document}
\maketitle

\begin{abstract}

Workflows are a fundamental component of automation in enterprise platforms, enabling the orchestration of tasks, data processing, and system integrations. Despite being widely used, building workflows can be complex, often requiring manual configuration through low-code platforms or visual programming tools. 
 To simplify this process, we explore the use of generative foundation models, particularly vision-language models (VLMs), to automatically generate structured workflows from visual inputs.
Translating hand-drawn sketches or computer-generated diagrams into executable workflows is challenging due to the ambiguity of free-form drawings, variations in diagram styles, and the difficulty of inferring execution logic from visual elements. 
To address this, we introduce \ourmodel{}, a framework for generating structured workflow outputs from sketches using vision-language models. We curate a diverse dataset of workflow diagrams—including synthetic, manually annotated, and real-world samples to enable robust training and evaluation. We finetune and benchmark multiple vision-language models, conducting a series of ablation studies to analyze the strengths and limitations of our approach. Our results show that finetuning significantly enhances structured workflow generation, outperforming large vision-language models on this task. All resources, including models, code, datasets, and evaluation metrics, are made available\footnote{\url{https://servicenow.github.io/StarFlow/}}.

\end{abstract}    
\section{Introduction}
\label{sec:intro}

Workflows play a crucial role in automating business processes, orchestrating data flows, and integrating enterprise applications. They enable organizations to streamline operations, reduce manual effort, and enforce business logic across complex systems \cite{mulesoftAutomation, servicenowFlowDesigner, microsoftPowerAutomate}. Despite their ubiquity, workflow creation remains a challenging task, often requiring users to manually configure processes through low-code platforms or visual programming environments. While these tools offer greater accessibility than traditional programming, they still demand a deep understanding of system logic, data dependencies, and execution rules. 

\begin{figure}[t]
    \centering
    \vspace{-12pt}
    \includegraphics[width=1.0\linewidth]{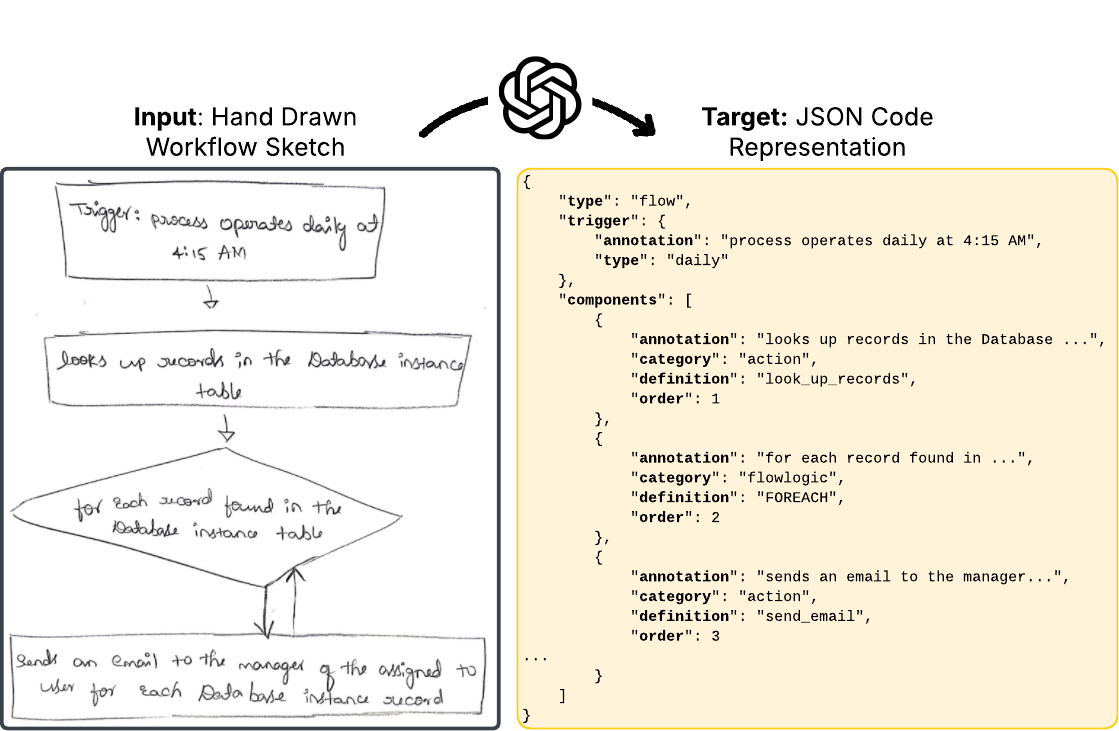}
    \vspace{-12pt}
    \caption{\textbf{The task of sketch to workflow.} Given an input image representing a business process, the task is to convert the logic of the diagram found in the image into a structured JSON output describing the execution logic of the workflow, including the appropriate trigger, actions, and inputs.}
    \label{fig:teaser}
    \vspace{-15pt}
\end{figure}

An intuitive alternative would be the ability to generate structured workflows directly from visual representations, such as hand-drawn sketches or diagrams, as portrayed in Figure \ref{fig:teaser}. However, this problem is inherently difficult due to the ambiguity of sketches, variations in diagramming conventions, and the complexity of extracting structured execution logic from visual elements.

In this work, we introduce \textbf\ourmodel{}, a framework designed to generate structured workflow representations from sketch-based inputs using vision-language models (VLMs). Our approach involves curating a diverse dataset comprising synthetic, manually annotated, and real-world workflow diagrams, which we use to finetune multiple vision-language models. To evaluate the performance of our approach, we use a \emph{flow similarity} metric that measures the structural fidelity of generated workflows based on the tree representation of the workflow and tree edit distance. Our results demonstrate that finetuning significantly enhances the ability of VLMs to generate structured workflows, outperforming general-purpose models on this specialized task. 


By addressing the limitations of existing workflow creation methods and demonstrating the effectiveness of vision-language models in this domain, our work represents a step toward making workflow automation more intuitive and accessible.
Our key contributions are as follows:

\begin{itemize}
    \vspace{-7pt}
    \item We introduce \ourmodel{}, a framework for converting hand-drawn and digital workflow sketches into structured representations, enabling seamless workflow automation.
    \vspace{-7pt}
    \item We build a diverse dataset of workflow diagrams, spanning synthetic, human-annotated, and real-world samples, to enhance training and evaluation.
    \vspace{-7pt}
\end{itemize}
To foster future research and reproducibility, we \textbf{open-source all resources}, including models and their implementations, code used for training, datasets, and evaluation metrics.

\section{Related Work}
\label{sec:related_work}

\vspace{-3pt}
\subsection{Structured Output and Code Generation}

Language models trained on code \citep[e.g.][]{chen2021evaluating, li2023starcoder, roziere2023code, hui2024qwen2, zhu2024deepseek} have seen significant advancements in recent years, improving various aspects of software development, including code generation~\cite{nijkamp2022codegen, jiang2024survey, rodriguez2024starvectorgeneratingscalablevector}, comprehension~\cite{feng2020codebert, lu2021codexglue}, and translation tasks~\cite{lachaux2020unsupervised, yan2023codetransocean}. These models utilize large-scale source code datasets~\cite[e.g.][]{kocetkov2022stack} to learn programming syntax and semantics, enabling them to produce functional and syntactically correct code snippets from natural language prompts.

Evaluation of code generation models is notoriously difficult. Metrics such as the HumanEval benchmark~\cite{chen2021evaluating} aim to evaluate a model's ability to generate functionally correct code solutions. Other metrics, such as CodeBLEU \cite{ren2020codebleu}, extend the traditional BLEU score~\cite{papineni2002bleu} by incorporating code-specific features such as syntax and data flow, offering a more nuanced evaluation of code generation quality. In this work, we draw inspiration from CodeBLEU and introduce a flow similarity metric based on tree representation and tree edit distance~\cite{zhang1989simple}.

\vspace{-4pt}
\subsection{Multimodal Large Language Models}
\vspace{-2pt}

Vision-language models (VLMs) \citep[e.g.][]{alayrac2022flamingo, liu2023visual, agrawal2024pixtral, wang2024qwen2, dubey2024llama} have made significant strides in integrating visual and textual data, enabling more sophisticated multimodal understanding. These models excel at various tasks, including image captioning~\cite{lin2014microsoft, vinyals2015show}, visual question answering~\cite{antol2015vqa, hudson2019gqa, yue2024mmmu}, and document understanding~\cite{rodriguez2024bigdocs, tong2025cambrian}.

One task that remains challenging for VLMs is to generate code or structured outputs based on a screenshot or diagram~\cite{liu-etal-2022-code, shukla2023towards, shi2025chartmimic, rodriguez2024bigdocs, herrera2017flow2code}. For example, \citet{shi2025chartmimic} introduce a benchmark to assess the performance of VLMs on generating code to reproduce charts. Closely related to our work, \citet{liu-etal-2022-code} propose a two-step process to generate code from a flowchart via two distinct models, one to extract the structure of the diagram, and the other to generate executable code from pseudocode. In this paper, we focus on generating structured workflows in JSON format from hand-drawn or computer generated sketches.

\vspace{-4pt}
\subsection{Workflow Generation}
\vspace{-2pt}

Recent work on workflow generation from textual inputs has demonstrated significant advancements in the field of automated task planning and execution. Approaches relying on retrieval-augmented generation and task decomposition have been shown to be effective for solving this problem~\cite{bechard2024reducing, bassamzadeh2024comparative, ayala2024generating}. Other notable efforts include \citet{zeng2023flowmind}, who built models to generate workflows for specific applications, \citet{fan2024workflowllm} who develop a synthetic data pipeline used to train a workflow generator, and \citet{cai2023low} who built a graphical user interface allowing a user to build and edit a workflow with the assistance of an LLM. In this work, we focus on generating workflows from hand-drawn sketches and computer generated diagrams instead of doing so from textual instructions.

\section{Methodology}
\label{sec:methodology}

In this section, we go over the dataset creation process and how we evaluate generated flows. We first present a quick overview of what workflows are. We then discuss how we build synthetic workflows by finding patterns frequently found in ones that appear in the real-world. Finally, we highlight how we programmatically create diagrams for these workflows, and how we use these samples as a basis for the human-annotated data.

\subsection{The Anatomy of a Workflow}

\textit{Workflows} are automated processes that consist of a sequence of reusable \textit{actions} that perform operations on a user's data. Within a workflow, actions are intertwined with \textit{flow logic} elements, such as conditions and loops, that control the execution of the workflow. A workflow typically includes a \textit{trigger} that determines when the execution starts. Alternatively, a \textit{subflow} consists of the same actions and flow logic as a workflow, but does not include a trigger. Subflows are meant to be called by workflows or other subflows, similar to how functions are used in programming languages. 

Workflows can be triggered in a variety of ways. For example, a workflow can start after a certain interval of time has passed, when a record has been updated in a given table, or when an email is received, to name a few.
The actions found in workflows can also perform a variety of operations on behalf of a user. For example, they can look up a set of records in a given table, make updates to records, send emails, connect to third-party APIs, and much more.

\subsection{Synthetic Workflow Generation}

Real world workflows are often built using a distinct set of design patterns. To build our synthetic workflow generation pipeline, we implemented a heuristic that can build workflows using a set of flow logic elements (e.g. IF, ELSE, FOREACH) along with actions and subflows sampled either deterministically or stochastically based on the pattern. Algorithm \ref{alg:scheduled_loop} presents a simplified look at the code used for creating a workflow following the \emph{Scheduled Loop} pattern, which performs actions on multiple records at predefined time intervals.

\begin{algorithm}
\captionsetup{font=10pt,labelfont={bf,10pt}}
\small
\caption{Pseudocode for Scheduled Loop pattern}
\label{alg:scheduled_loop}
\begin{algorithmic}[1]
    \State Select a random Scheduled trigger
    \State Add a Look Up Records action for a table
    \State Add a \textbf{FOREACH} flow logic
    \If{random() $<$ $P_{IF}$} 
        \State Add an \textbf{IF} flow logic
    \EndIf
    \State Select a random action related to the table
    \If{IF statement exists and random() $<$ $P_{IF}$} 
        \State Add an \textbf{ELSE} flow logic
        \State Select another related action
    \EndIf
\end{algorithmic}
\end{algorithm}

After creating the workflows, we generate natural language annotations for each step using a large language model --- in our case, we used Llama 3.1 70B Instruct \cite{dubey2024llama}. We represent the resulting workflows in JSON format, which serves as the generation target for the VLM. Figure \ref{fig:scheduled_loop_json} in Appendix \ref{sec:appendix_json_workflow} presents an example flow generated using the \textit{Scheduled Loop} heuristic.

Once the synthetic workflows are generated, we proceed to creating variants of these samples using a variety of methods, thus obtaining workflow diagrams of five different flavors: \textsc{Synthetic}, \textsc{Manual}, \textsc{Digital}, \textsc{Whiteboard}, and \textsc{User Interface}. We describe the generation process of each in the next section.

\subsection{Creating Workflow Diagrams}

\textsc{Synthetic} workflows are created by programmatically generating a graph representation of each workflow using Graphviz~\cite{ellson2002graphviz}, including random variations in graph orientation and edge style. For example, the graph representation of the workflow in Figure~\ref{fig:scheduled_loop_json} is shown in 
Figure \ref{fig:scheduled_loop_graph} (Appendix \ref{sec:appendix_graph_workflow}).

To create the \textsc{User Interface} workflows, we further render the programmatically generated flows using {ServiceNow}'s native visualization tool, as illustrated in Figure \ref{fig:scheduled_loop_ui} (Appendix \ref{sec:appendix_ui_workflow}). This offers an alternative representation of the flows within an environment that closely aligns with potential deployment scenarios.

The three workflow types \textsc{Manual}, \textsc{Digital}, and \textsc{Whiteboard} are created by human annotators. We contracted an external vendor to recruit annotators to create flow diagrams based on the synthetically generated ones. The annotators were given these graph representations and were asked to create flow diagrams for each graph sample using either digital tools (\textsc{Digital}), or by drawing the graph on paper (\textsc{Manual}) or on a whiteboard or blackboard (\textsc{Whiteboard}). Details regarding the human annotators can be found in Appendix \ref{sec:appendix_annotators}. Figure \ref{fig:scheduled_loop_sketch} in Appendix \ref{sec:appendix_sketch_workflow} presents such an example for the same flow found in Figure \ref{fig:scheduled_loop_json}.

For each flow JSON in our dataset, we generate one or more images using the approach described above. We then divide the samples according to the flow JSON, ensuring that no flows are shared between the different dataset splits. The number of samples generated for each sample type can be found in Table \ref{tab:dataset_splits}. 

\begin{table}[ht]
    \vspace{-5pt}
    \centering
    \small
    \begin{tabular}{lccc}
        \hline
        \textbf{Source} & \textbf{Train} & \textbf{Valid} & \textbf{Test} \\
        \hline
        \textsc{Synthetic}   & 12,376 & 1,000  & 1,000  \\
        \textsc{Manual}      & 3,035  & 333    & 865    \\
        \textsc{Digital}     & 2,613  & 241    & 701    \\
        \textsc{Whiteboard}  & 484    & 40     & 46     \\
        \textsc{User Interface} & 373   & 116    & 87     \\
        \hline
        Total & 18,881 & 1,730 & 2,699 \\
        \hline
    \end{tabular}
    \caption{\textbf{Dataset distribution across splits}. Samples are collected from a variety of sources, ranging from visualizations generated synthetically to hand-drawn samples. Examples for each type of sample can be found in Appendix \ref{sec:appendix_sample_types}.}
    \label{tab:dataset_splits}
    \vspace{-20pt}
\end{table}

\section{Experiments}
\label{sec:experiments}

In this section, we conduct experiments to assess the capabilities of various open-weight and proprietary VLM models on the Sketch-to-Workflow task and its evaluation metrics. Additionally, we examine whether finetuning improves performance on the downstream task.

\vspace{-5pt}
\subsection{Models} \label{sec:models}

We perform our experiments using a variety of frontier models as well as open-weight alternatives. We evaluate the following proprietary models: GPT-4o and GPT-4o-mini \cite{hurst2024gpt},  Claude-3.7-Sonnet \cite{claude3-7-sonnet-card}, Gemini-2.0-Flash \cite{team2023gemini}. We put these models head-to-head against a set of strong open-weights alternatives, namely Pixtral \cite{agrawal2024pixtral}, LLaMA 3.2 Vision (11B and 90B) \cite{dubey2024llama},  Phi-3.5 \cite{abdin2024phi}, and Qwen2.5-VL (3B, 7B and 72B) \cite{bai2025qwen2}. Additionally, we finetune the smaller variants of the open-weight models and observe the resulting improvements on downstream tasks. Training details for the finetuned models can be found in Appendix \ref{sec:appendix_training}.

\subsection{Evaluation of Generated Workflows}
Assessing the quality of generated flows presents challenges similar to those in evaluating generated code. In this work, we report four types of metrics that provide a comprehensive evaluation by capturing different aspects of flow generation. The metrics we report are Flow Similarity ($FlowSim$), Tree BLEU ($TreeBLEU$), Trigger Match ($TM$), and Component Match ($CM$). 
%
For \emph{Flow Similarity}, we follow the methodology used in \citet{ayala2024generating}: we decompose generated workflows into trees and compute the tree edit distance using the algorithm from \citet{zhang1989simple}. We normalize the obtained tree edit distance by the number of nodes in each tree to obtain a score between 0 and 1. 

\vspace{-18pt}
\begin{equation}\label{eq:flow_sim}
    \text{FlowSim}(F, F_r) = 1 - \frac{TED(F, F_r)}{|F| + |F_r|}
\end{equation}
\vspace{-12pt}

\noindent where $F$, $F_r$ denote the given flow and the reference flow, respectively.

We use a custom weighting scheme that assigns greater weight to changes affecting actions than those affecting inputs. Figure \ref{fig:scheduled_loop_tree} in Appendix \ref{sec:appendix_tree_workflow} illustrates the tree decomposition derived from the flow JSON defined in Figure \ref{fig:scheduled_loop_json}.

We also use a variant of \emph{TreeBLEU}~\cite{gui2025webcode2m} that leverages our tree decomposition to assess structural hierarchy recall between flows. 

\vspace{-12pt}
\begin{equation}\label{eq:tree_bleu}
    \text{TreeBLEU}(F, F_r) = \frac{|S(F)\cap S(F_r)|}{|S(F)|}
\end{equation}
\vspace{-12pt}

\noindent where $S(.)$ denotes the set of 1-height subtrees.

To ensure fairness, we exclude subtrees of height 1 that are always present --- specifically, the \texttt{Flow $\rightarrow$ Trigger} and \texttt{Flow $\rightarrow$ Components} edges --- so that empty flows without triggers or components receive a score of zero.

\emph{Trigger Match} measures the percentage of cases where the model correctly predicts the trigger from the sample. \emph{Component Match}, on the other hand, computes the intersection between the predicted and target components, normalized by their union. This metric evaluates the model’s ability to predict the correct components in an order-agnostic manner, akin to the \textit{bag-of-components} metric from \cite{bechard2024reducing}. Equation \ref{eq:match_metrics} depicts both the Trigger Match and Component Match metrics.

\vspace{-15pt}
\begin{equation}\label{eq:match_metrics}
    \text{TM} = \mathbf{1}_{\{T_F = T_{F_r}\}}
    \qquad
    \text{CM} = \frac{|C_F \cap C_{F_r}|}{|C_F \cup C_{F_r}|}
\end{equation}
\vspace{-12pt}

\noindent where $T_F$ and $T_{F_r}$ denote the trigger of the given and reference flows, and $C_F$ and $C_{F_R}$ denote the set of components in each flow.

\begin{table*}[ht]
    \centering
    \footnotesize
    \tabcolsep 2pt
    \begin{tabular}{lcc|cc|cc}
        \hline
         & \textbf{FlowSim} & \textbf{FlowSim} & \textbf{TreeBLEU} & \textbf{TreeBLEU} & \textbf{Trigger} & \textbf{Component} \\
        \textbf{Model} & \textcolor{gray}{\small{w/ inputs}} & \textcolor{gray}{\small{no inputs}} & \textcolor{gray}{\small{w/ inputs}} & \textcolor{gray}{\small{no inputs}} & {\small{\textcolor{gray}{match}}} & {\small{\textcolor{gray}{match}}}\\
        \hline
        \multicolumn{7}{c}{\textit{\textbf{Open-weights Models}}} \\
        \rowcolor{open_models_below_4B!50}
        Qwen-2.5-VL-3B-Instruct~\citep{wang2024qwen2} & \textbf{0.410} & \textbf{0.384} & \textbf{0.360} & \textbf{0.329} & 0.027 & \textbf{0.201} \\
        \rowcolor{open_models_below_4B!50}
        Phi-3.5-Vision-4B-Instruct\citep{abdin2024phi} & 0.364 & 0.346 & 0.337 & 0.295 & \textbf{0.079} & 0.193 \\
        \hline
        \rowcolor{open_models_below_12B!50}
        Phi-4-Multimodal-6B-Instruct~\citep{abouelenin2025phi} & 0.465 & 0.404 & 0.394 & 0.298 & 0.054 & 0.244 \\
        \rowcolor{open_models_below_12B!50}
        Qwen-2.5-VL-7B-Instruct~\citep{wang2024qwen2} & \underline{0.614} & \underline{0.538} & \underline{0.562} & \underline{0.508} & 0.036 & \textbf{0.280} \\
        \rowcolor{open_models_below_12B!50}
        LLaMA-3.2-11B-Vision- Instruct~\citep{dubey2024llama} & 0.466 & 0.435 & 0.416 & 0.382 & \underline{0.075} & 0.239 \\
        \rowcolor{open_models_below_12B!50}
        Pixtral-12B~\citep{agrawal2024pixtral} & \textbf{0.632} & \textbf{0.582} & \textbf{0.617} & \textbf{0.541} & \textbf{0.088} & 0.261 \\
        \hline
        \rowcolor{open_models_over_12B!50}
        Qwen-2.5-VL-72B- Instruct~\citep{wang2024qwen2} & \textbf{0.710} & \textbf{0.643} & \textbf{0.703} & \textbf{0.655} & 0.325 & \textbf{0.305} \\
        \rowcolor{open_models_over_12B!50}
        LLaMA-3.2-90B-Vision-Instruct~\citep{dubey2024llama} & 0.687 & 0.603 & 0.681 & 0.627 & \textbf{0.328} & 0.286 \\

        \hline
        \multicolumn{7}{c}{\textit{\textbf{Proprietary Models}}} \\
        \rowcolor{closed_models!70}
        GPT-4o-Mini~\citep{hurst2024gpt} & 0.642 & 0.617 & 0.650 & 0.623 & 0.254 & 0.305 \\
        \rowcolor{closed_models!70}
        GPT-4o~\citep{hurst2024gpt} & \textbf{0.786} & \underline{0.707} & \underline{0.794} & \underline{0.718} & 0.282 & \underline{0.317} \\
        \rowcolor{closed_models!70}
        Claude-3.7-Sonnet~\citep{claude3-7-sonnet-card} & 0.763 & 0.679 & 0.769 & 0.701 & \underline{0.318} & 0.305 \\
        \rowcolor{closed_models!70}
        Gemini Flash 2.0~\citep{team2023gemini} & \underline{0.780} & \textbf{0.713} & \textbf{0.798} & \textbf{0.743} & \textbf{0.466} & \textbf{0.329} \\
        \hline
        \multicolumn{7}{c}{\textit{\textbf{Finetuned Models}}} \\
        \rowcolor{open_models_below_4B!50}
        Qwen-2.5-VL-3B-Instruct~\citep{wang2024qwen2} & \textbf{0.941} & \textbf{0.911} & \textbf{0.941} & \textbf{0.902} & \textbf{0.775} & \textbf{0.909} \\
        \rowcolor{open_models_below_4B!50}
        Phi-3.5-Vision-4B-Instruct~\citep{abdin2024phi} & 0.917 & 0.882 & 0.917 & 0.869 & 0.703 & 0.874 \\
        \hline
        \rowcolor{open_models_below_12B!50}
        Phi-4-Multimodal-6B-Instruct~\citep{abouelenin2025phi} & 0.939 & 0.908 & 0.940 & 0.901 & 0.770 & 0.907 \\
        \rowcolor{open_models_below_12B!50}
        Qwen-2.5-VL-7B-Instruct~\citep{wang2024qwen2} & \textbf{0.957} & \textbf{0.927} & \textbf{0.956} & \textbf{0.920} & \textbf{0.819} & \textbf{0.934} \\
        \rowcolor{open_models_below_12B!50}
        LLaMA-3.2-11B-Vision-Instruct~\citep{dubey2024llama} & \underline{0.955} & \underline{0.924} & \underline{0.954} & \underline{0.915} & \underline{0.805} & \textbf{0.934} \\
        \rowcolor{open_models_below_12B!50}
        Pixtral-12B~\citep{agrawal2024pixtral} & 0.952 & 0.919 & 0.950 & 0.908 & 0.753 & 0.930 \\
        \hline
    \end{tabular}
    \vspace{-6pt}

        \caption{\textbf{Flow quality metrics comparison across different models}. We compare proprietary models against open-weight models and their finetuned versions, with higher values indicating better performance for each metric. The best metric within each model category is highlighted in \textbf{bold}, and the runner up is \underline{underlined} if there are more than two models in that category. Models are categorized by size: \colorbox{open_models_below_4B!50}{blue for models smaller than 4B parameters}, \colorbox{open_models_below_12B!50}{orange for models between 4B and 12B parameters}, \colorbox{open_models_over_12B!50}{green for models larger than 12B parameters}, and \colorbox{closed_models!70}{gray for proprietary models}.
        We evaluate models by making a single workflow generation call to the language model. 
        } 
    
    \label{tab:flow_similarity}
    \vspace{-14pt}
\end{table*}

\subsection{Sketch to Workflow} \label{sec:sketch2flow}

In this section, we assess the performance of models listed in Section \ref{sec:models} on the task of sketch to workflow generation. 
We evaluate models that are proprietary and ones that have open weights, across a variety of model sizes. Our experiments indicate that (1) most proprietary models perform better than open-weights ones without any domain-specific training, and that (2) finetuning on \ourmodel{} helps open-weights models outperform proprietary models. 
Our results are summarized in Table \ref{tab:flow_similarity}.

Our results show that most proprietary models perform well on the workflow generation task. As expected, GPT-4o-mini underperforms compared to larger models, likely due to its smaller size.
Among open-weight models, all models from the Qwen2.5-VL family of models perform remarkably well against models of similar sizes. Pixtral is another strong model for its size, nearly matching the performance of the larger Llama variant.
In addition, performance trends remain consistent across different evaluation metrics. Across all models, scores for FlowSim and TreeBLEU are closely aligned, whether or not input conditions are considered. Additionally, finetuned models perform strongly on the trigger match TM metric, whereas proprietary and non-finetuned open-weight models lag further behind.

Finetuning significantly improves performance, surpassing all baselines by a substantial margin. In particular, the finetuned version of Qwen-2.5-VL-7B achieves notably high scores compared to all other models, closely followed by Llama 3.2 11B and Pixtral-12B.
We hypothesize that finetuned models acquire crucial domain knowledge during training, which proprietary models struggle to replicate without additional external information. For example, when prompted with an image representing a flow for creating a user in Microsoft Azure Active Directory, a proprietary model must infer the type, definition, and scope of the relevant component.  If the model predicts a component of type \texttt{action} with definition name \texttt{create\_user} in scope \texttt{ms\_azure\_active\_directory}, but the actual answer is a component of type \texttt{action} with definition name \texttt{create\_a\_user} in scope \texttt{sn\_ms\_ad\_spoke}, it receives a score of zero.

Finetuned models benefit from exposure to such components during training, allowing them to memorize proper naming conventions of different components and improve accuracy.
There are several potential ways to mitigate this issue. One approach is to integrate tool calls, enabling the VLM to retrieve relevant components during generation. Another is to incorporate retrieval-augmented generation (RAG) by extracting relevant details directly from images. Alternatively, breaking down the task into smaller subtasks could facilitate more effective retrieval of contextual information, helping to ground VLMs during generation.

\begin{figure*}[ht]
    \centering
    \includegraphics[width=\linewidth]{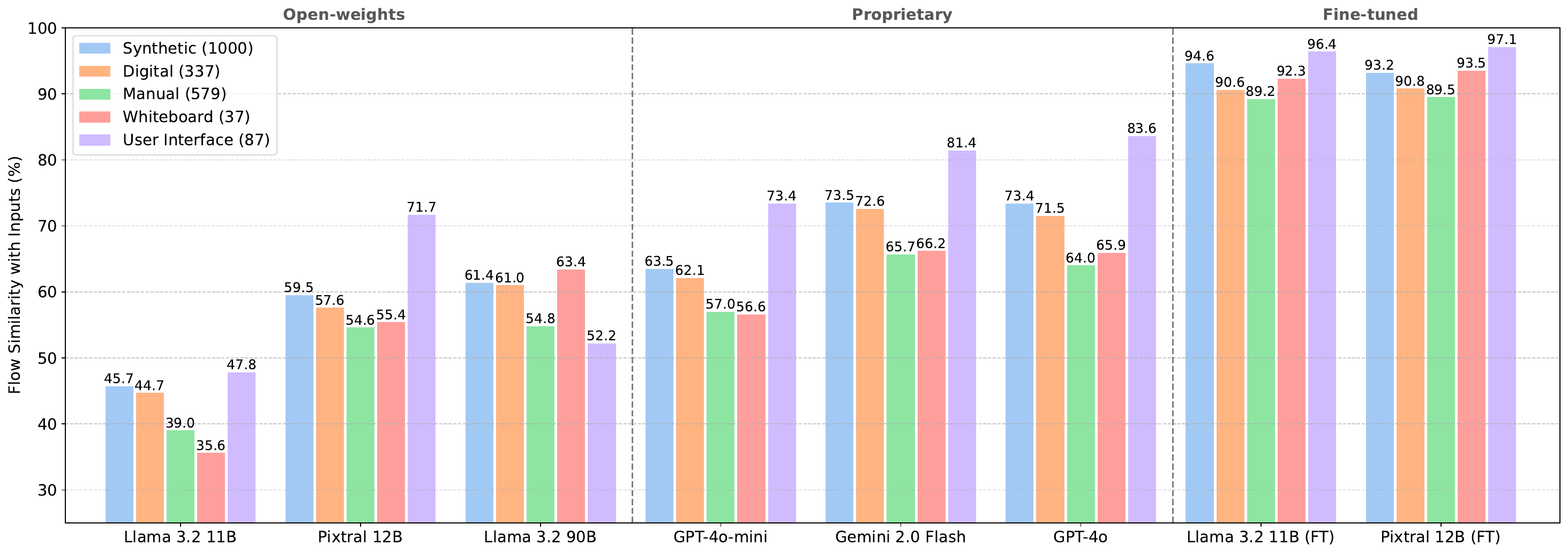}
    \vspace{-20pt}
    \caption{\textbf{Performance of each model per type of sample}. We report the FlowSim with inputs results. Number of supporting examples for each sample type is shown in parenthesis. Examples for each type of sample can be found in Appendix \ref{sec:appendix_sample_types}.}
    \label{fig:sample_source}
    \vspace{-15pt}
\end{figure*}

\begin{figure}[ht]
    \vspace{-5pt}
    \centering
    \includegraphics[width=\linewidth]{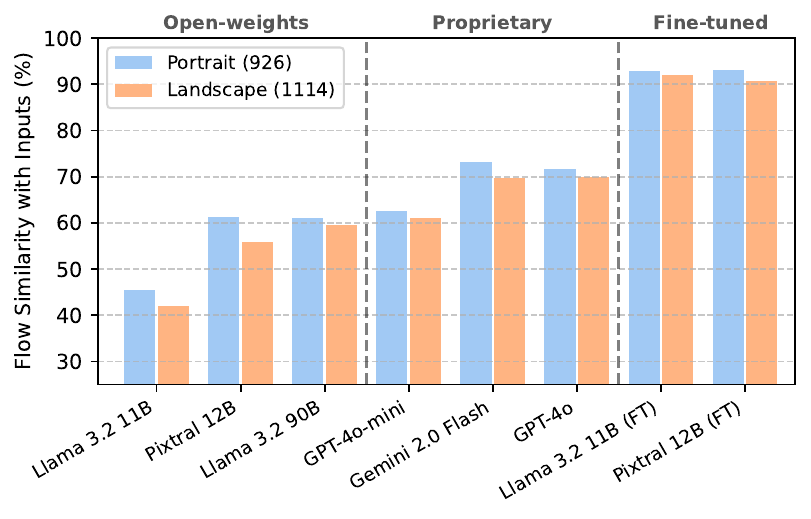}
    \vspace{-26pt}
    \caption{\textbf{Impact of image orientation}. We report the FlowSim with inputs results. Number of supporting examples for each sample type is shown in parenthesis.}
    \label{fig:sample_orientation}
\vspace{-10pt}
\end{figure}

\subsection{Evaluation by Subpopulation}

We are interested in understanding whether some models have more difficulty generating flows for certain types of images. We perform our analysis amongst three distinct axes: source of sample, orientation of sample, and image size. We perform these experiments on a subset of the models described in Section \ref{sec:models}.

\subsubsection{Source of Sample}\label{sec:sample_source}

In section \ref{sec:experiments}, we observed that models finetuned with \ourmodel{} generally outperform ones that are not finetuned in workflow generation. One question is whether these models perform better across all types of samples.
To answer this question, we evaluate a subset of the models discussed in Section \ref{sec:sketch2flow} on a stratified version of our dataset. Results are shown in Figure~\ref{fig:sample_source}.

\begin{figure}[ht]
    \vspace{-5pt}
    \centering
    \small
    \includegraphics[width=\linewidth]{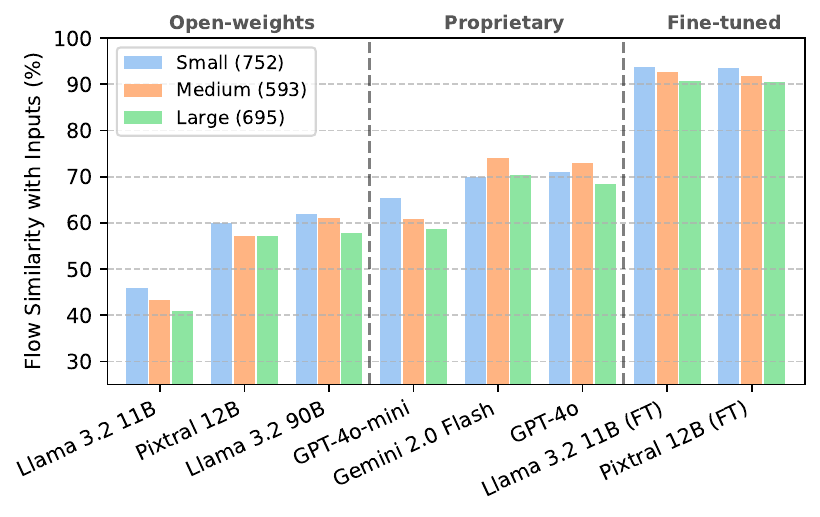}
    \vspace{-20pt}
    \caption{\textbf{Impact of image resolution}. We report the FlowSim \textit{with input} results. The number of supporting examples for each sample type is shown in parentheses.}
    \label{fig:image_size}
    \vspace{-10pt}
\end{figure}

We find that all models experience a drop in performance on the \textsc{Manual} samples compared to other types of samples, closely followed by \textsc{Whiteboard} samples. Intuitively, these images are harder to interpret as they require the model to read handwritten text in order to properly select components used to generate the workflow. On the other hand, we find that \textsc{User Interface} screenshots and \textsc{Synthetic} samples are the easiest samples. Since these samples are rendered automatically, we hypothesize that they contain the least amount of ambiguity regarding the execution logic of the workflow. \textsc{User Interface} samples contain more textual information than other types, as the interface interprets the flow and presents additional details about the triggers and components (see Appendix \ref{sec:appendix_ui_workflow}). This extra context can make the task easier for models.

\vspace{-2pt}
\subsubsection{Orientation of Sample}\label{sec:sample_orientation}

Workflow diagrams can be represented horizontally or vertically without changing meaning. As such, we are interested in assessing whether models are better at interpreting sketches presented top-to-bottom (portrait) versus left-to-right (landscape). Our criteria for differentiating the two sample types is based on the aspect ratio of the image. We define samples where images are twice as wide as tall as \textit{landscape}, and the rest as \textit{portrait}.

Our results, summarized in Figure \ref{fig:sample_orientation}, show that all benchmarked models exhibit a slight drop in performance and that this gap is more pronounced for the non-finetuned variant of Pixtral-12B (even as this gap is reduced after finetuning). We hypothesize that part of the difference in performance might be explained by the composition of each split. For example, \textsc{User Interface} samples, which are easier (see Section \ref{sec:sample_source}), are largely portrait samples due to the nature of the data collection. The presence of such samples in the Portrait category may partially explain the performance gap.

\subsubsection{Ablation on Image Resolution}

We study the effect of image resolution on model performance. We split sample images into three categories based on size: small (less than 400k pixels), large (more than 1M pixels), and medium (in between). We choose these image sizes as boundaries to ensure categories are approximately the same size. We present results in Figure \ref{fig:image_size}.

We observe that GPT-4o and Gemini-2.0-Flash perform better on samples of medium size compared to smaller or larger samples by a non-negligible margin. This trend does not repeat for open-weight models other models as they both perform better in smaller images. It is worth noting that Pixtral's performance remains stable for larger images, whereas LLaMA exhibits a consistent degradation as image size increases. Finally, we observe this trend of degraded performance on larger samples remaining after finetuning, although it is to a lesser extent.

\subsection{Generalization Beyond Training Distribution} \label{sec:generalization}

We are interested in the capability of our models to generalize to out-of-distribution settings. Specifically, we fine-tuned Llama 3.2 11B Vision Instruct and Pixtral 12B exclusively on synthetically generated workflow diagrams (\textsc{Synthetic}) and evaluated their performance across diverse and out-of-distribution (OOD) diagram styles (see Table~\ref{tab:sample_type_generalization}). 

\begin{table}[ht]
    \centering
    \small
    \tabcolsep 5pt
    \begin{tabular}{lcccccc}
        \hline
        & \textbf{\textsc{Synt}} & \textbf{\textsc{Digi}} & \textbf{\textsc{Man}} & \textbf{\textsc{WB}} & \textbf{\textsc{UI}} & \textbf{Avg} \\
        \hline
        \multicolumn{7}{c}{\textit{\textbf{Llama 3.2 11B Vision Instruct}}} \\
        Base & 45.7 & 44.7 & 39.0 & 35.6 & 47.8 & 43.5 \\
        FT (Synth) & \underline{92.9} & \underline{65.9} & \underline{66.7} & \underline{53.0} & \underline{58.1} & \underline{78.7} \\
        FT (All) & \textbf{94.3} & \textbf{90.0} & \textbf{88.3} & \textbf{91.0} & \textbf{95.3} & \textbf{91.9} \\
        \hline
        \multicolumn{7}{c}{\textit{\textbf{Pixtral 12B}}} \\
        Base & 59.5 & 57.6 & 54.6 & 55.4 & \underline{71.7} & 58.8 \\
        FT (Synth) & \underline{92.1} & \underline{86.0} & \underline{80.0} & \underline{79.2} & 58.3 & \underline{86.0} \\
        FT (All) & \textbf{92.7} & \textbf{90.8} & \textbf{89.4} & \textbf{92.4} & \textbf{97.0} & \textbf{91.6} \\
        \hline
    \end{tabular}
    \vspace{-5pt}
    \caption{\textbf{Performance of models by sample type (\%)}. Models finetuned on solely on synthetic data outperform the base models in most cases. The reported metric is Flow Similarity with Inputs.}
    \label{tab:sample_type_generalization}
    \vspace{-7pt}
\end{table}

\begin{figure}[ht]
    \centering
    \vspace{-5pt}
    \includegraphics[width=\linewidth]{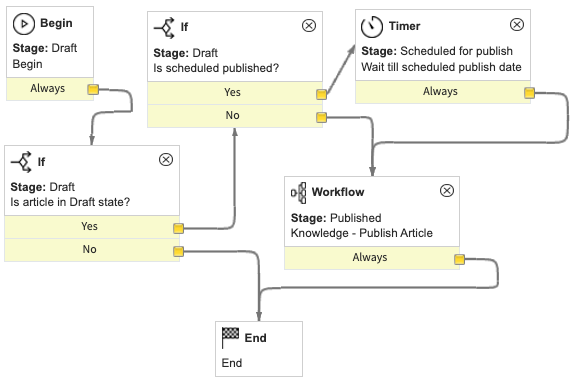}
    \vspace{-20pt}
    \caption{\textbf{Screenshot of a workflow}. The screenshot was taken from a visualization platform for which we don't have examples in the training dataset.}
    \label{fig:workflow}
    \vspace{-10pt}
\end{figure}

Results show that models trained solely on synthetic data achieve large gains over their base versions: for Llama 3.2 11B Vision Instruct, the average Flow Similarity increases from 43.5 (base) to 78.7, and for Pixtral 12B from 58.8 to 86.0. Nevertheless, fine-tuning on the full, diverse dataset yields the best results, reaching 91.9 for Llama and 91.6 for Pixtral, highlighting the continued importance of data diversity.

\begin{figure}[ht]
    \centering
    \includegraphics[width=\linewidth]{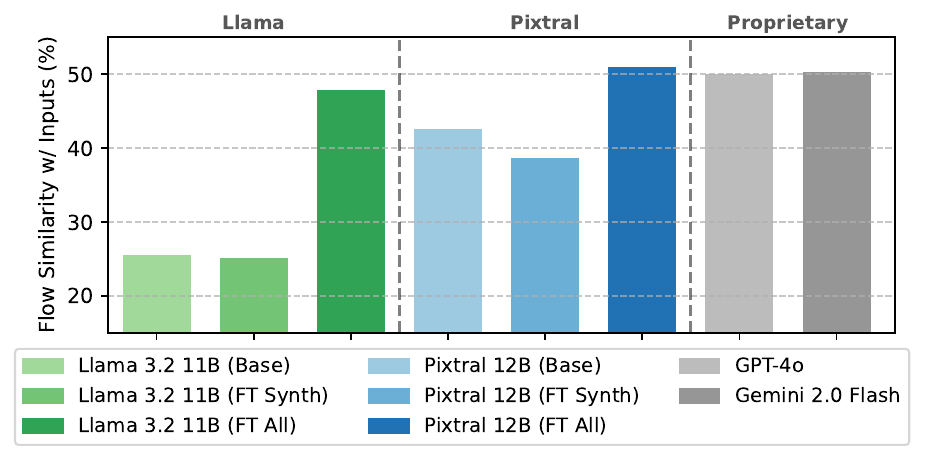}
    \vspace{-20pt}
    \caption{\textbf{Out-of-distribution generalization results.} Models finetuned on diverse data perform on par with proprietary models.}
    \label{fig:ood_generalization}
    \vspace{-18pt}
\end{figure}

Moreover, to better assess generalization, we evaluate models on 300 OOD samples, incorporating real-world and human-generated diagrams that vary in style and complexity. Figure \ref{fig:workflow} shows an example, originating from a workflow platform not represented in the training data. The OOD evaluation set reveals that our fine-tuned models substantially outperform their base counterparts, while fine-tuned Pixtral even surpasses GPT-4o and Gemini (see Figure~\ref{fig:ood_generalization}). These results validate the effectiveness of our approach and highlight potential improvements with more data diversity. We also note that performance across all models is much lower than on the main dataset, indicating room for growth in quality.

\begin{figure*}[t]
    \centering
    \vspace{-10pt}
    \begin{subfigure}{0.24\textwidth}
        \centering
        \includegraphics[width=\textwidth]{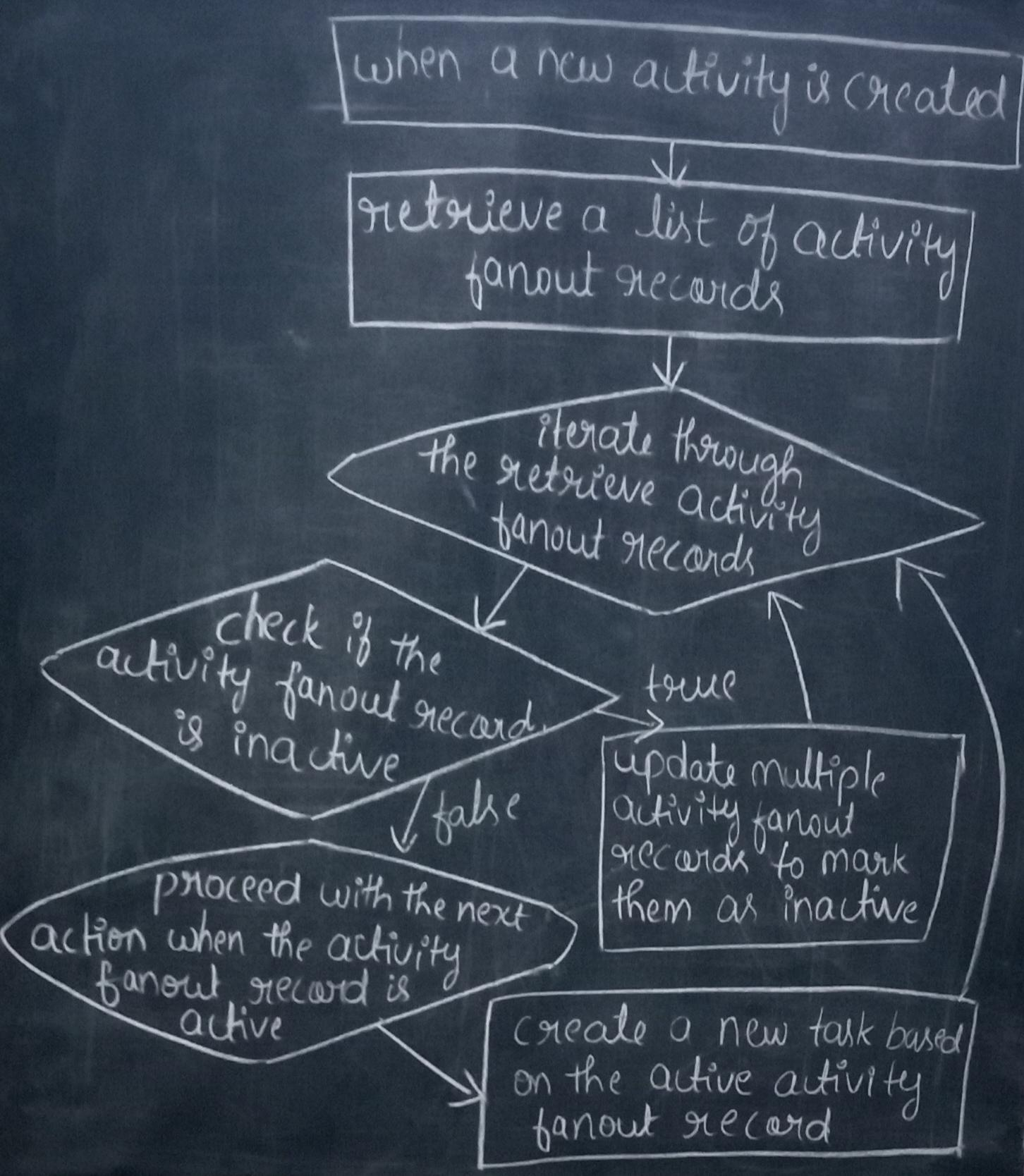}
        \caption{Ground Truth}
        \label{fig:running_example_truth}
    \end{subfigure}
    \begin{subfigure}{0.24\textwidth}
        \centering
        \includegraphics[width=\textwidth]{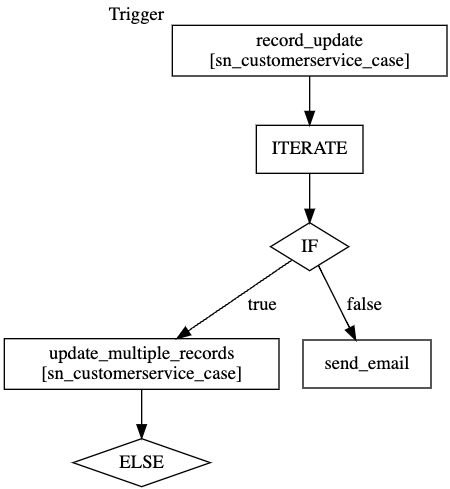}
        \caption{Llama 3.2 11B}
        \label{fig:running_example_llama_base}
    \end{subfigure}
    \begin{subfigure}{0.24\textwidth}
        \centering
        \includegraphics[width=\textwidth]{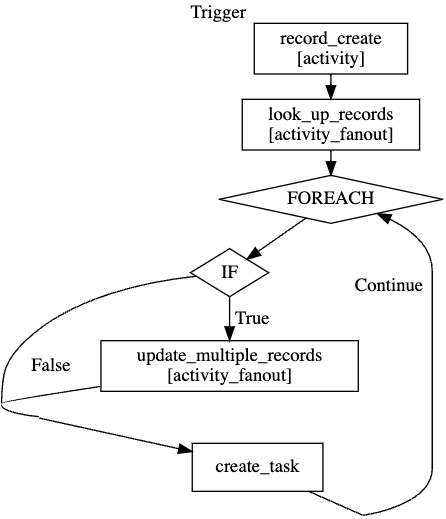}
        \caption{GPT-4o}
        \label{fig:running_example_gpt}
    \end{subfigure}
    \begin{subfigure}{0.24\textwidth}
        \centering
        \includegraphics[width=\textwidth]{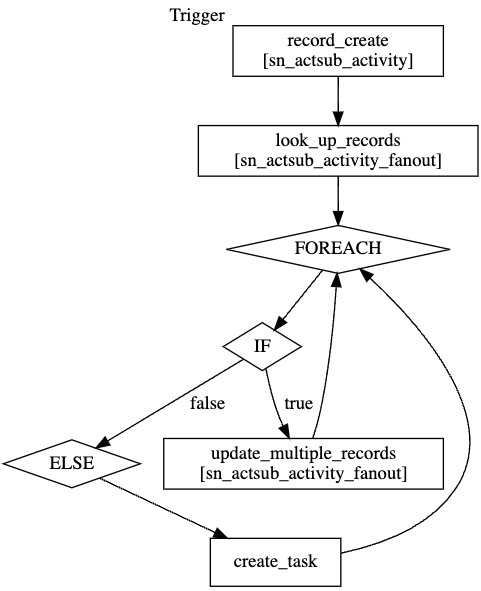}
        \caption{Llama 3.2 11B Finetuned}
        \label{fig:running_example_llama_ft}
    \end{subfigure}
    \caption{Blackboard sketch of a workflow along with rendered workflows generated by different VLMs.}
    \label{fig:running_example}
\end{figure*}

\section{Error Analysis and Discussion}
\label{sec:results}

In this section, we examine the current failure modes of various models in workflow generation. To illustrate this, we present a representative example that highlights the strengths and limitations of each approach. We will use the flow depicted in Figure \ref{fig:running_example_truth} to compare the capabilities of each model qualitatively. For the sake of brevity, we will focus on Llama 3.2 11B, a finetuned variant of that same model, and GPT-4o.

When prompting a non-finetuned Llama 3.2 11B model to generate a flow, the model can struggle with several basic failures, such as predicting the wrong trigger for the task, and picking an unrelated table. Moreover, the model fails to use flowlogic elements properly, and hallucinates actions unrelated to the sketch, such as adding a component to send an email. Resulting flow can be seen in Figure \ref{fig:running_example_llama_base}

A strong proprietary model like GPT-4o does perform qualitatively better on the task. In Figure \ref{fig:running_example_gpt}, we observe that the model is able to properly predict the trigger and most of the components without generating unrelated ones. However, we see that the model occasionally struggles with keeping track of the flow execution logic, where it omits an \texttt{ELSE} statement in the flow. The model also encounters difficulties with some fine-grained details in the flow that pertain to the component inputs, such as selecting a generic \texttt{activity} table when unsure of the appropriate name. In the above example, we see that the model falls back to using a generic \texttt{activity} table when unsure about what the name of the table should be given the provided information.

Finally, the finetuned variant of Llama 3.2 11B performs better than its counterparts on this example. The model predicts the appropriate flow execution logic along with all relevant components. It is also able to properly predict the right tables for the task as it has seen some data from the same domain during the finetuning phase. We include more qualitative examples illustrating limitations of our approach on various diagram styles, and compare outputs from multiple base and fine-tuned models in Appendix \ref{sec:more_error_analysis}.
\section{Conclusion}
\label{sec:conclusion}

In this paper, we presented \ourmodel{}, a framework for structured workflow generation from sketch-based diagrams. By leveraging vision-language models and a diverse dataset, we demonstrated that finetuned models outperform general-purpose models at accurately translating sketches into structured workflow representations. Our experiments revealed key insights into the challenges posed by different sketch sources, orientations, and image resolutions, highlighting the importance of domain-specific training.

While our approach shows strong performance in workflow generation, future work could explore extending the methodology to a broader range of workflow visualization styles and improving robustness to handwritten annotations. Additionally, refining evaluation metrics to consider functional execution correctness could provide a more comprehensive assessment of generated workflows. Finally, augmenting models with external information via retrieval-augmented generation or function calling might help better ground the models in generating accurate information in the workflows. Overall, \ourmodel{} represents a step toward making workflow automation more accessible and intuitive by enabling seamless sketch-to-workflow generation.
\section*{Limitations}
\label{sec:limitations}

While \textsc{StarFlow} demonstrates strong performance in translating workflow sketches into structured outputs, several limitations remain. First, our models rely heavily on the diversity and fidelity of the training dataset. Although we curated synthetic, human-drawn, and interface-based diagrams, the coverage of real-world workflow styles and enterprise-specific notations is still incomplete, which may affect generalization to unseen diagram conventions. This issue is amplified in out-of-distribution scenarios: as shown in our OOD experiments, diagrams that diverge substantially from the training distribution (in layout, annotation style, or component positioning) lead to increased structural and semantic errors.

Second, evaluation is primarily based on structural similarity metrics (FlowSim, TreeBLEU, Trigger/Component Match). These metrics assess syntactic alignment but do not capture whether the generated workflow would execute correctly or satisfy user intent. Further, the current evaluation protocol does not measure the frequency or severity of hallucinated components. Future work should quantify hallucinations explicitly and incorporate execution-based or behavior-level evaluation (e.g. as seen in \cite{bechard2024reducing, bassamzadeh2024comparative}).

Third, despite substantial improvements from finetuning, the models still struggle to ground generation to real-time information. For example, verifying whether a referenced component or table actually exists in the environment or documentation remains a challenge. This limitation sometimes leads to the creation of invalid or unused components. Stronger grounding mechanisms (e.g., retrieval-augmented generation with access to API schemas or component registries) could help mitigate this.

Fourth, our approach remains sensitive to noisy inputs such as messy handwriting, cluttered diagrams, or ambiguous component names, suggesting a need for more robust integrating of visual parsing with semantic priors or external knowledge. Finally, finetuning large vision–language models incurs substantial computational cost, limiting accessibility in low-resource settings.

Future work could address these challenges by expanding dataset coverage with more heterogeneous, real-world samples, incorporating execution-based evaluation metrics, explicitly tracking component hallucination rates, and exploring retrieval-augmented or tool-assisted generation to improve grounding, generalization, and computational efficiency.

\paragraph{Compute \& variance.} Due to compute constraints, some results are from a single seed; we therefore provide scripts to reproduce runs. Future work will widen the seed sweep and include confidence intervals.

{
    \small
    \bibliography{main}
}

\clearpage
\appendix
\section{Example JSON from the Workflow Generation Heuristic}\label{sec:appendix_json_workflow}

\begin{figure}[H]
\begin{lstlisting}[language=json, lineskip=-0.1pt]
{
  "type": "flow",
  "scope": "global",
  "trigger": {
    "annotation": "on wednesdays at a quarter to 5 pm",
    "type": "weekly",
    "inputs": [
      {
        "name": "day_of_week",
        "value": "3"
      },
      {
        "name": "time",
        "value": "1970-01-01 16:45:00"
      }
    ]
  },
  "components": [
    {
      "annotation": "look up incident tasks",
      "category": "action",
      "definition": "look_up_records",
      "scope": "global",
      "order": 1,
      "inputs": [
        {
          "name": "table",
          "value": "incident_task"
        }
      ]
    },
    {
      "annotation": "for all",
      "category": "flowlogic",
      "definition": "FOREACH",
      "scope": "global",
      "order": 2,
      "inputs": [
        {
          "name": "items",
          "value": "{{1.Records}}"
        }
      ]
    },
    {
      "annotation": "if the task is inactive",
      "category": "flowlogic",
      "definition": "IF",
      "scope": "global",
      "order": 3,
      "block": 2,
      "inputs": [
        {
          "name": "condition",
          "value": "{{2.item.active}}=false"
        }
      ]
    },
    {
      "annotation": "post incident details on MS Teams",
      "category": "action",
      "definition": "post_incident_details",
      "scope": "sn_ms_teams_ah",
      "order": 4,
      "block": 3
    }
  ]
}
\end{lstlisting}
\caption{\textbf{Example of a flow generated by the Scheduled Loop heuristic.} The flow includes a scheduled trigger, lookup action, conditional logic, and an MS Teams action.}
\label{fig:scheduled_loop_json}
\end{figure}
\section{End-to-End vs Task Decomposition}

Here, we compare whether our end-to-end baseline approach of sketch to workflow generation can match the performance of a pipeline that decomposes the task into multiple subtasks. Following a methodology that closely matches \cite{ayala2024generating}, we first introduce the task of sketch to workflow summary, which aims to boil down the different actions performed in a given workflow sketch into a natural language summary. Then, we use this summary to first generate a workflow outline from the generated summary, and finally generate inputs for the trigger and each action found in the generated flow outline iteratively. This modular approach allows us to use a more sophisticated approach to sketch to workflow generation that can incorporate search calls to retrieve relevant actions and inputs to include in the final flow. For each of our experiments, we use GPT-4o as the image summarizer and a different model for workflow generation (proprietary, open-weights, or finetuned). Results are presented in Table \ref{tab:task_decomposition}.

\begin{table}[ht]
    \centering
    \small
    \begin{tabular}{lcc}
        \hline
        & \textbf{FlowSim} & \textbf{FlowSim} \\
        \textbf{Model}& \textcolor{gray}{\small{no input}} & \textcolor{gray}{\small{w/ input}} \\
        \hline
        \multicolumn{3}{c}{\textit{\textbf{Sketch $\rightarrow$ Workflow}}} \\
        GPT-4o & \textit{0.786} & \textit{0.707} \\
        Pixtral-12B & 0.632 & 0.582 \\
        Pixtral-12B (ft) & \textbf{0.952} & \textbf{0.919} \\
        \hline
        \multicolumn{3}{c}{\textit{\textbf{Sketch $\rightarrow$ Summary $\rightarrow$ Outline ${\rightarrow}$ Workflow}}} \\
        GPT-4o & 0.727 & 0.647 \\
        Mistral-Nemo-Instruct-2407 & 0.472 & 0.414 \\
        Mistral-Nemo-Instruct-2407 (ft) & \textbf{0.834} & \textbf{0.828} \\
        \hline
    \end{tabular}
    \caption{\textbf{Impact of Task Decomposition on Flow Similarity.} Flow similarity scores for the end-to-end Sketch to Flow task compared to a stepwise approach that decomposes the task into subtasks (Sketch $\rightarrow$ Summary $\rightarrow$ Flow Outline $\rightarrow$ Flow with Inputs).}    
    \label{tab:task_decomposition}
\end{table}

We find that decomposing the task into multiple subtasks yields lower results across the tested models. This is most likely due to errors compounding every step of the generation pipeline: every small detail missed by the summarization step will impact the generation of the flow outline, which will itself impact which inputs get populated for each component. Moreover, keeping the task of image to flow generation as a single task can substantially decrease the total latency of the application as the number of total calls to the LLM or VLM are significantly reduced.

\section{Types of Workflow Diagram Samples}
 \label{sec:appendix_sample_types}

\subsection{Example of Synthetic Flow Graph}\label{sec:appendix_graph_workflow}

\begin{figure}[H]
    \centering
    \includegraphics[width=0.8\linewidth]{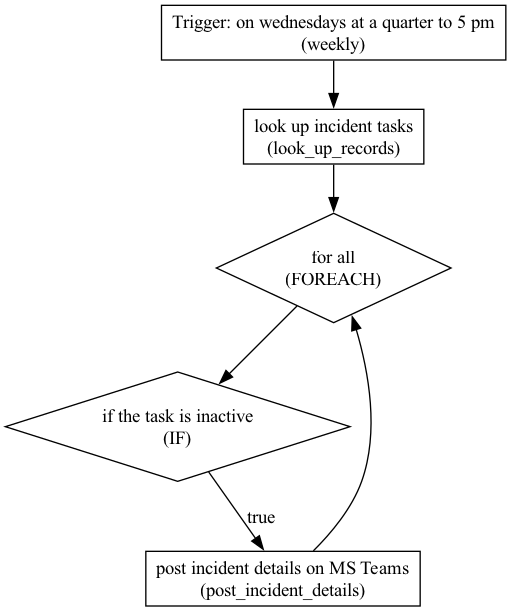}
    \caption{Graph-based representation of a workflow generated programmatically (Synthetic sample).}
    \label{fig:scheduled_loop_graph}
\end{figure}

\subsection{Example of Digital Flow Sketch}

\begin{figure}[H]
    \centering
    \includegraphics[width=0.8\linewidth]{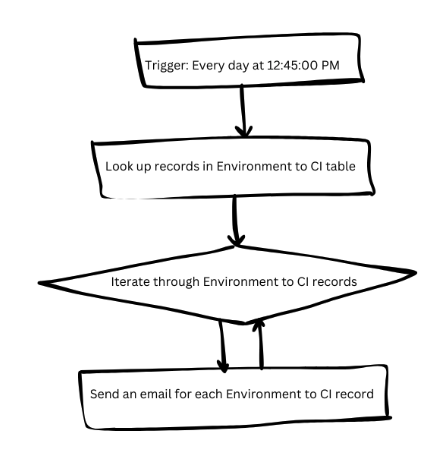}
    \caption{Digitally drawn representation of a workflow (Digital sample).}
    \label{fig:scheduled_loop_digital_sketch}
\end{figure}

\subsection{Example of Manual Flow Sketch} \label{sec:appendix_sketch_workflow}

\begin{figure}[H]
    \vspace{-5pt}
    \centering
    \includegraphics[width=0.8\linewidth]{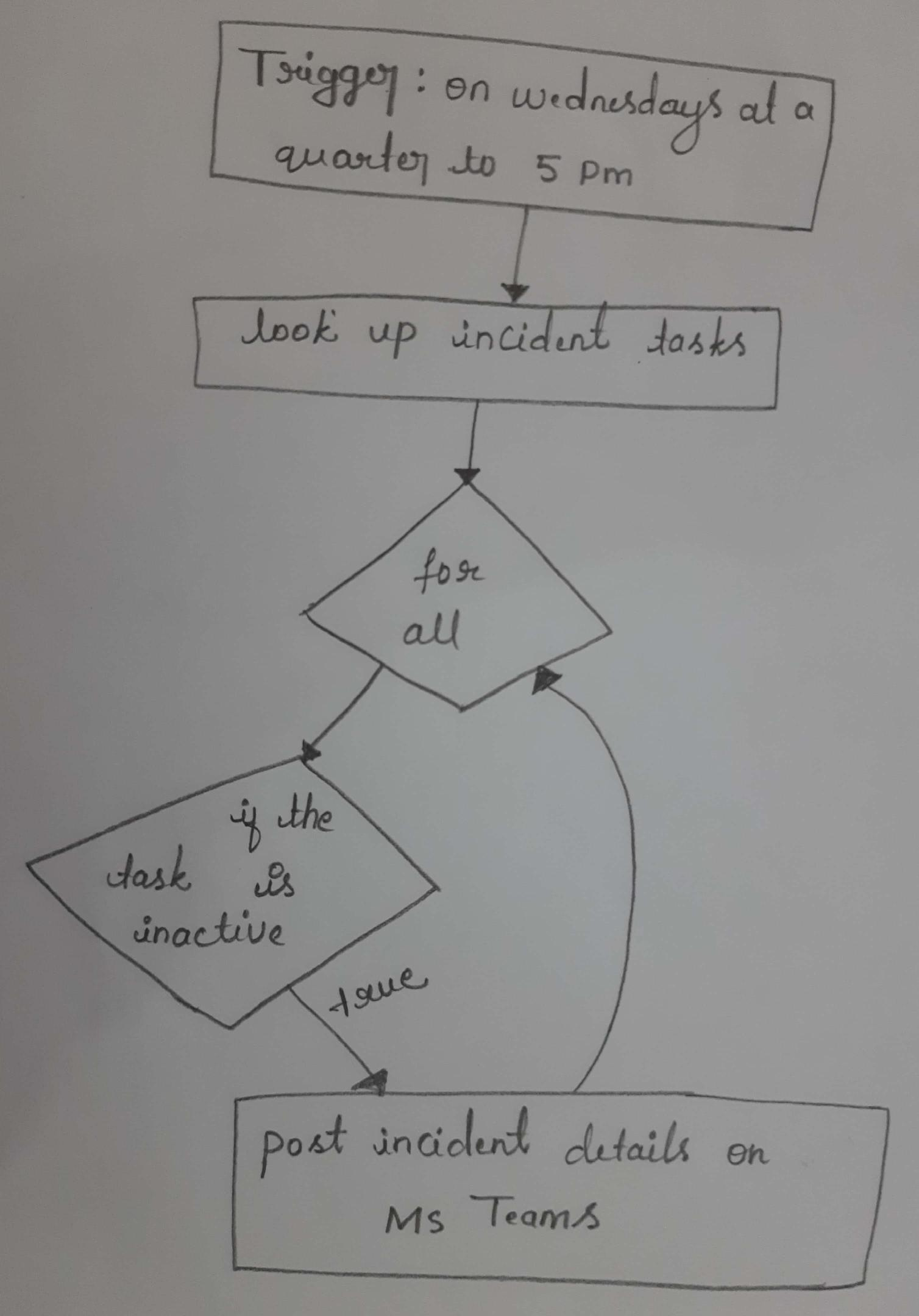}
    \caption{Hand-drawn representation of a workflow on paper (Manual sample).}
    \label{fig:scheduled_loop_sketch}
    \vspace{-5pt}
\end{figure}

\subsection{Example of Whiteboard Flow Sketch}

\begin{figure}[H]
    \centering
    \includegraphics[width=0.8\linewidth]{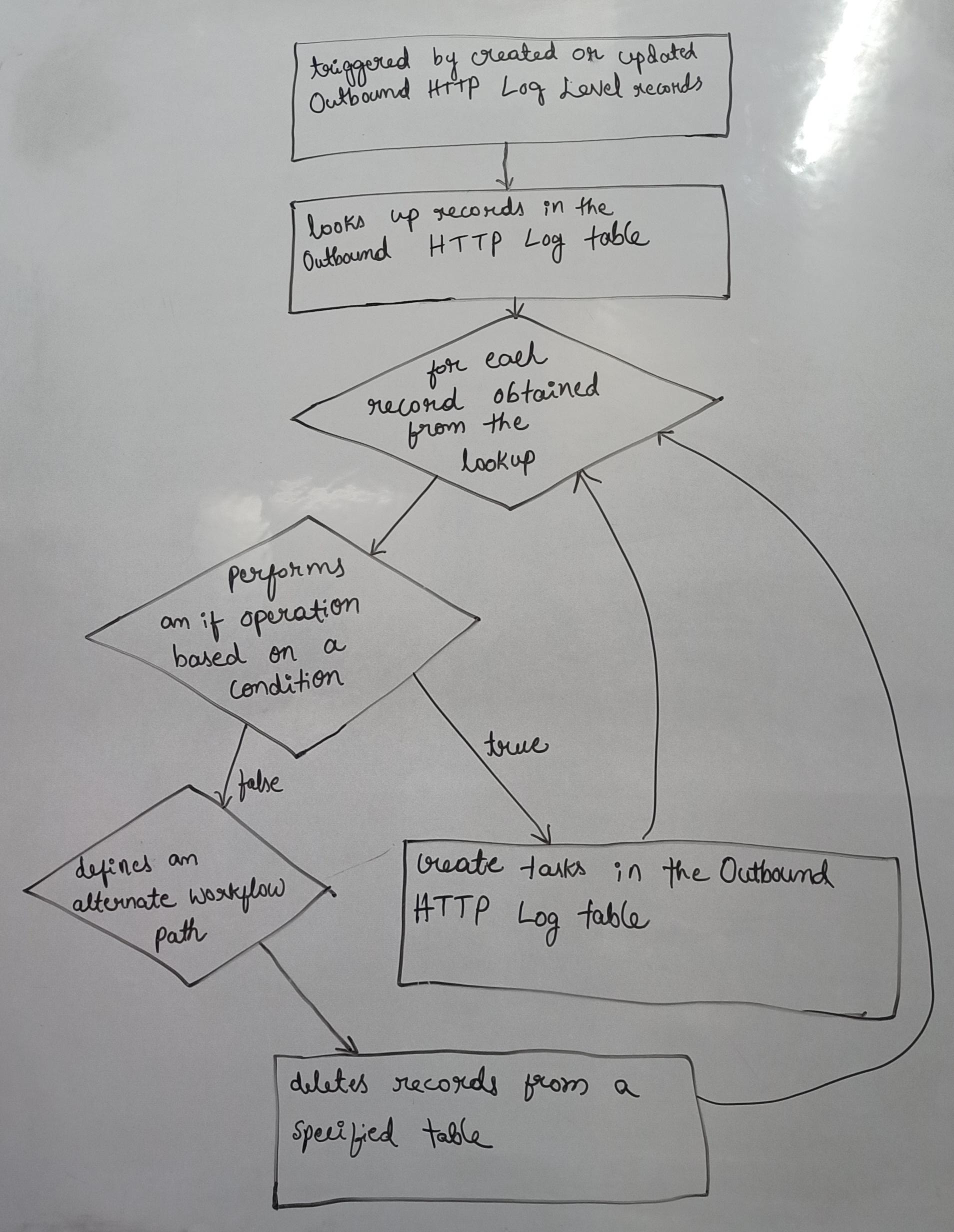}
    \caption{Hand-drawn representation of a workflow on a whiteboard (Whiteboard sample).}
    \label{fig:output_whiteboard}
\end{figure}

\subsection{Example of Workflow Rendered in a User Interface} \label{sec:appendix_ui_workflow}

\begin{figure}[H]
    \centering
    \includegraphics[width=0.8\linewidth]{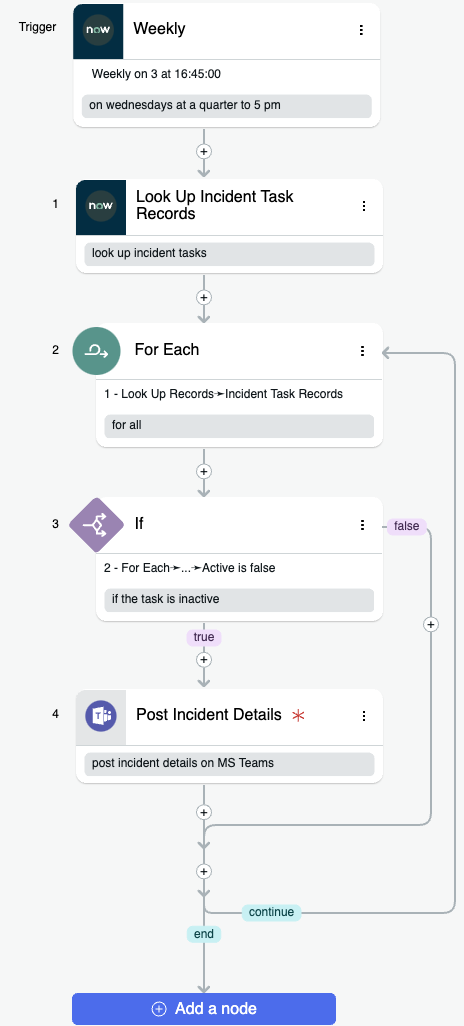}
    \caption{Rendered workflow as displayed in the ServiceNow User Interface (User Interface sample).}
    \label{fig:scheduled_loop_ui}
\end{figure}
\section{Human Annotators} \label{sec:appendix_annotators}

We partnered with a for-profit data labeling company (referred to as the "Vendor") specializing in data curation for AI applications. The annotation process spanned a three-month period, beginning with a pilot phase in the first month. During this phase, we collaborated closely with the Vendor's annotation team, conducting detailed reviews and providing extensive feedback to ensure annotators fully understood the task requirements.

Our dataset was annotated by a dedicated team of 24 professionals based in India. These annotators possessed strong proficiency in technical writing and English, with educational backgrounds primarily in engineering, computer science, and related disciplines. The majority held bachelor's degrees, while some had advanced degrees in specialized fields. Additionally, they brought prior experience in data labeling, ensuring familiarity with structured annotation tasks.

To ensure the highest standards of annotation quality, a comprehensive quality assurance framework was implemented, requiring each annotation to undergo at least three independent review stages. The process began with an initial annotation conducted by experienced annotators or trainers, followed by a primary quality assurance review, where a specialist assessed accuracy, completeness, and adherence to annotation guidelines. Finally, a secondary review ensured consistency and alignment with evolving project requirements. This structured, multi-tiered approach reinforced annotation quality, minimized inconsistencies, and enhanced dataset reliability.

To uphold ethical labor standards and maintain high annotation quality, all annotators were compensated at rates exceeding fair market wages in their respective countries. This strategy supports the recruitment and retention of highly skilled professionals, fostering long-term engagement and ensuring annotation consistency across the project.

\paragraph{Content Safety.} Annotators were instructed not to include any real names, emails, IDs, or screenshots of live systems. All hand-drawn images depict synthetic entities. A three-stage QA (initial review, primary QA, secondary QA) rejected any samples containing potentially identifying text. For UI-rendered samples, flows were generated from synthetic seeds only. We ran an OCR pass to spot obvious PII tokens prior to packaging.
\section{Training Details} \label{sec:appendix_training}

We use a consistent training setup for all finetuned models presented in this paper. To mitigate overfitting, we applied early stopping based on evaluation loss. The learning rate was initialized at $2 \times 10^{-5}$, and we used the AdamW optimizer \cite{loshchilov2017decoupled} with $\beta$ values of $(0.95, 0.999)$, weight decay of $1 \times 10^{-6}$, and an epsilon value of $1 \times 10^{-8}$ to ensure numerical stability. The learning rate followed a cosine schedule with a warmup phase of 30 steps. Additionally, we enforced a maximum gradient norm of 1.0 to prevent gradient explosion.

For all finetuning runs, we trained both the language model and the connector components of the VLM while keeping the vision encoder frozen. Each model was trained to support sequences of up to 32k tokens, including both image and text inputs.

We conducted training using 16 NVIDIA H100 80GB GPUs across two nodes. Full Sharded Data Parallel (FSDP) \cite{rajbhandari2020zero} was employed without CPU offloading. We also used mixed-precision training with bfloat16 (\texttt{bf16}).

\section{Decomposition of a Workflow into its Tree Representation} \label{sec:appendix_tree_workflow}

\begin{figure}[H]
    \centering
    \includegraphics[width=\linewidth]{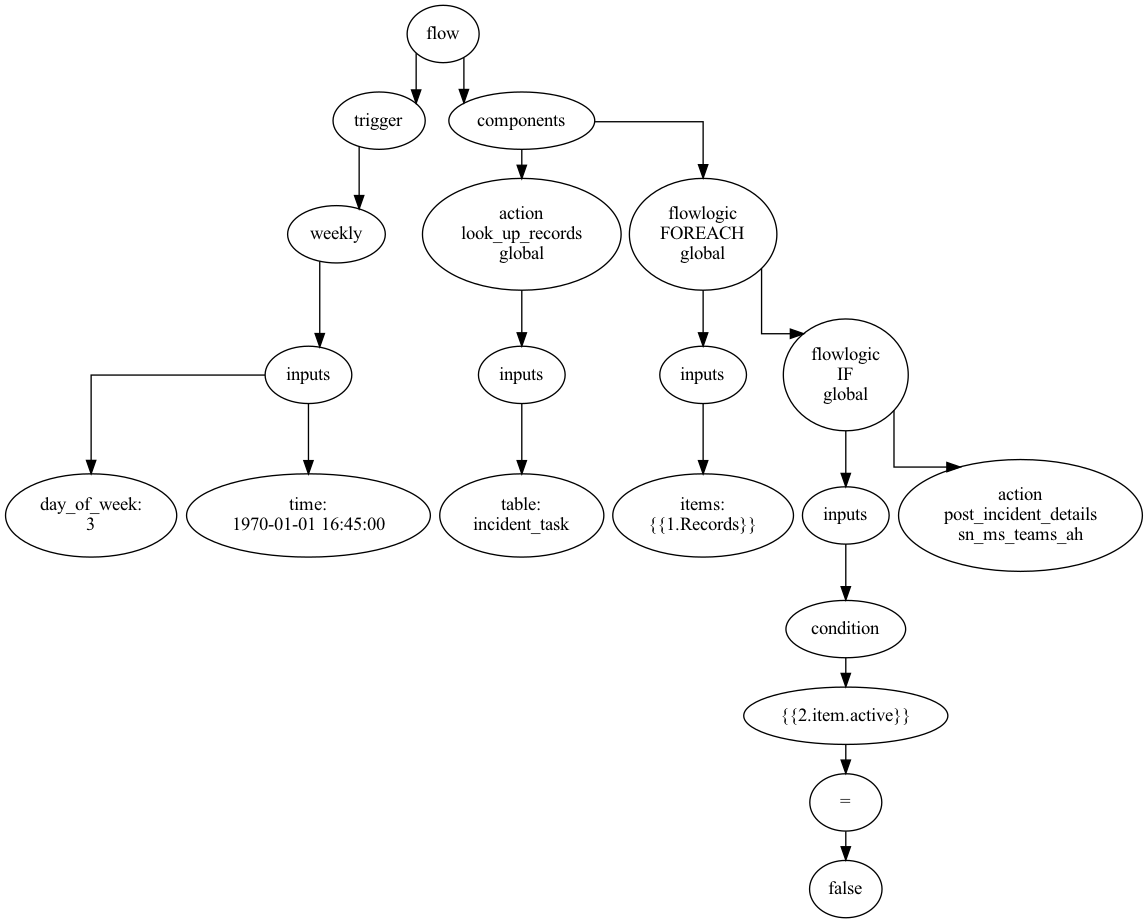}
    \caption{Decomposition of a workflow into its tree structure. This representation is used to compute Flow Similarity and TreeBLEU metrics.}
    \label{fig:scheduled_loop_tree}
\end{figure}
\section{Potential Risks}

While \ourmodel{} offers a promising step toward automating workflow generation, it is crucial to emphasize that automatically generated workflows should never be executed directly in production environments. Generated flows may contain logical errors, unsafe configurations, or unintended actions due to model hallucinations or misinterpretation of visual inputs.

To mitigate these risks, all workflows must be reviewed and verified by human experts prior to deployment. We strongly recommend that models be integrated within sandboxed or staging environments that allow thorough functional testing and validation of every action and trigger. Automated safeguards should prevent execution on live systems without explicit human approval.

In future iterations, we plan to include automated static and dynamic validation checks to detect potentially unsafe or destructive actions before execution, further reinforcing the human-in-the-loop principle that underpins responsible workflow automation.
\section{Ethical Statement}

Automatically generated workflows may contain logical errors, unsafe configurations, or unintended actions due to model misinterpretation or hallucination. All generated workflows must be reviewed by humans and exercised first in a sandbox/staging environment with guardrails that block destructive operations (e.g., writes to production tables, external API calls). We recommend static checks (schema/permission validation) and dynamic tests (dry-runs with synthetic data) prior to any deployment. Execution in live systems should require explicit human approval.

\paragraph{Artifact Use Consistency.} All external artifacts (open-weight VLMs and libraries) were used within their stated research licenses/terms. For \ourmodel{} artifacts, we specify research-only intended use; derivatives must comply with the original access conditions and must not be deployed in production automations without human oversight and sandbox validation.

\paragraph{Licensing \& Intended Use.}
The \ourmodel{} dataset and code are released under the Apache-2.0 license.
Finetuned model checkpoints inherit the license and usage terms of their respective base models; users must comply with those upstream licenses when using or redistributing our finetuned weights.

\paragraph{Use of AI Assistants.} We used AI assistants solely for wording/grammar suggestions on the manuscript draft; no content, code, data, or analysis was generated without human verification.
\section{Dataset and Model Cards}

We include a dataset card (domains, sources, splits, diagram types, known failure modes) and model cards (training data scope, objective, hyperparameters, evaluation metrics \& caveats, intended use/limits) alongside the artifacts. Everything can be found on the paper's website at \url{https://servicenow.github.io/StarFlow/}.

\section{Synthetic Dataset Details}
\label{app:dataset}

For completeness, we provide additional information on the dataset used for model training and evaluation. Each workflow instance was generated based on a predefined pattern template reflecting common enterprise automation scenarios. Table~\ref{tab:pattern_counts} presents the distribution of samples across all pattern types. This breakdown provides insight into the structural diversity of the dataset and serves as a reference for reproducibility.

\begin{table}[ht]
    \centering
    \small
    \begin{tabular}{lc}
        \hline
        \textbf{Pattern} & \textbf{Count} \\
        \hline
        \textsc{Crud Loop} & 2,148 \\
        \textsc{Crud Single} & 2,144 \\
        \textsc{Service Catalog Request Manual} & 1,102 \\
        \textsc{Scheduled Loop} & 1,100 \\
        \textsc{Scheduled Single} & 1,100 \\
        \textsc{Outbound Notification} & 1,100 \\
        \textsc{Integration Inbound} & 1,058 \\
        \textsc{Integration Batch Sync} & 1,000 \\
        \textsc{Single Component} & 994 \\
        \textsc{Sla} & 660 \\
        \textsc{Parallel} & 656 \\
        \textsc{Trigger Only} & 624 \\
        \textsc{Misc} & 364 \\
        \textsc{Pad} & 152 \\
        \textsc{Inbound Email New} & 60 \\
        \textsc{Service Catalog Request Automated} & 58 \\
        \textsc{Inbound Email Reply} & 56 \\
        \hline
        Total & 14,376 \\
        \hline
    \end{tabular}
    \caption{\textbf{Pattern Distribution}. A detailed breakdown of counts across different pattern types.}
    \label{tab:pattern_counts}
\end{table}

\section{Future Directions}

Building on the limitations identified above, several promising research directions emerge. First, future work should explore execution-based evaluation of generated workflows (e.g. \cite{chen2021evaluating, austin2021program}. While current metrics assess structural similarity, they do not capture whether the synthesized workflow functions as intended during runtime. Integrating simulation or sandboxed execution environments would enable semantic validation of control flow, component compatibility, and real-world task completion.

Second, agentic workflow generation represents a natural evolution beyond static sketch translation. Rather than producing workflows in a single forward pass, models could be embedded within reasoning-enabled frameworks (e.g., ReAct-style agents \cite{yao2022react}) capable of tool invocation \cite{schick2023toolformer, li2025start}, and self-correction \cite{gehring2024rlef}. Such agents could query documentation, perform component lookup, verify field or table availability, and incorporate real-time constraints before finalizing the generated flow. This hybrid approach would improve grounding and reduce hallucination.

Overall, these directions shift workflow generation from static prediction toward \textbf{interactive, execution-aware, and tool-augmented agents}, enabling robust automation and iterative co-design of flows with human operators.
\section{More Error Analysis}\label{sec:more_error_analysis}

\begin{figure*}[p]
    \centering

    \begin{subfigure}{0.48\textwidth}
        \centering
        \includegraphics[width=0.9\linewidth]{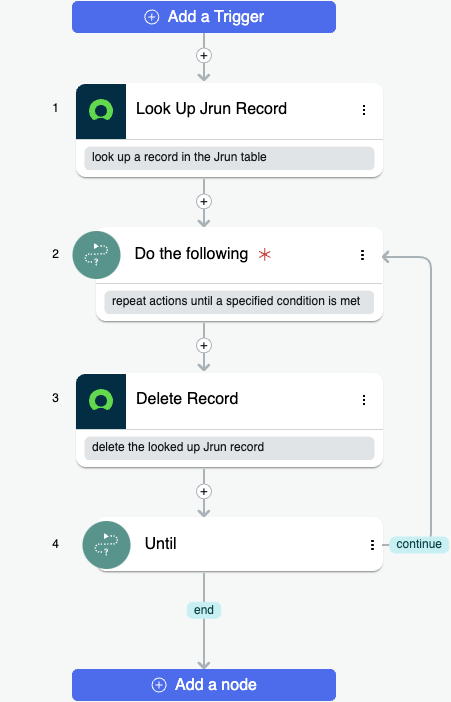}
        \caption{\textsc{User Interface} Workflow Diagram}
        \label{fig:sub1}
    \end{subfigure}
    \hfill
    \begin{subfigure}{0.48\textwidth}
        \centering
        \includegraphics[width=0.7\linewidth]{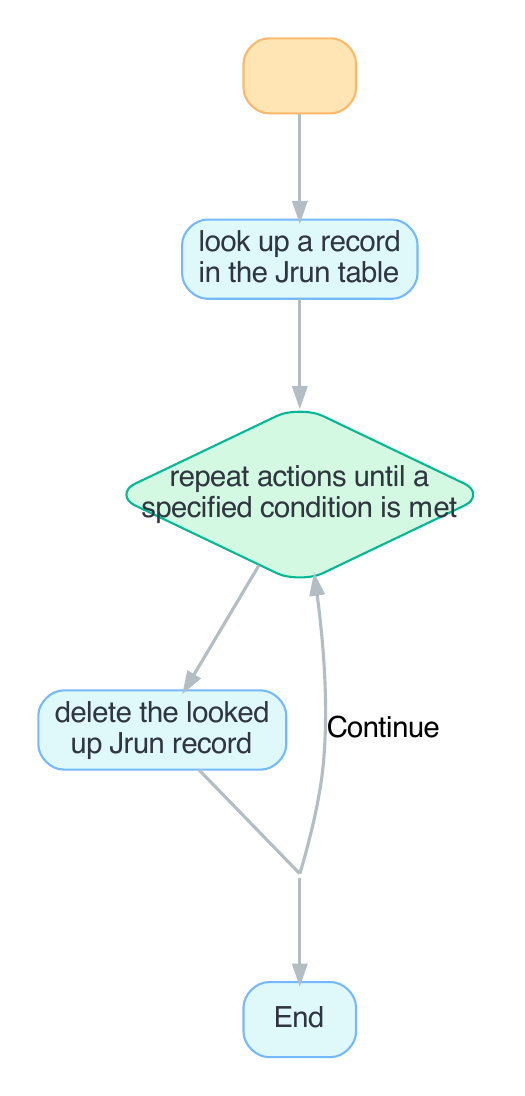}
        \caption{GPT-4o}
        \label{fig:sub2}
    \end{subfigure}

    \vspace{1cm}

    \begin{subfigure}{0.48\textwidth}
        \centering
        \includegraphics[width=0.8\linewidth]{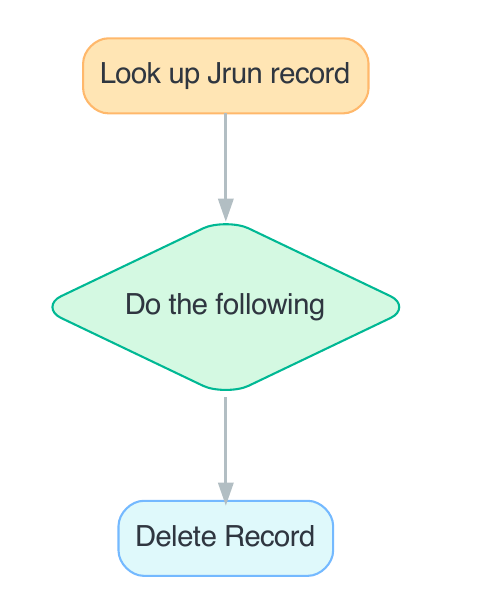}
        \caption{Qwen-2.5-VL-7B-Instruct}
        \label{fig:sub3}
    \end{subfigure}
    \hfill
    \begin{subfigure}{0.48\textwidth}
        \centering
        \includegraphics[width=\linewidth]{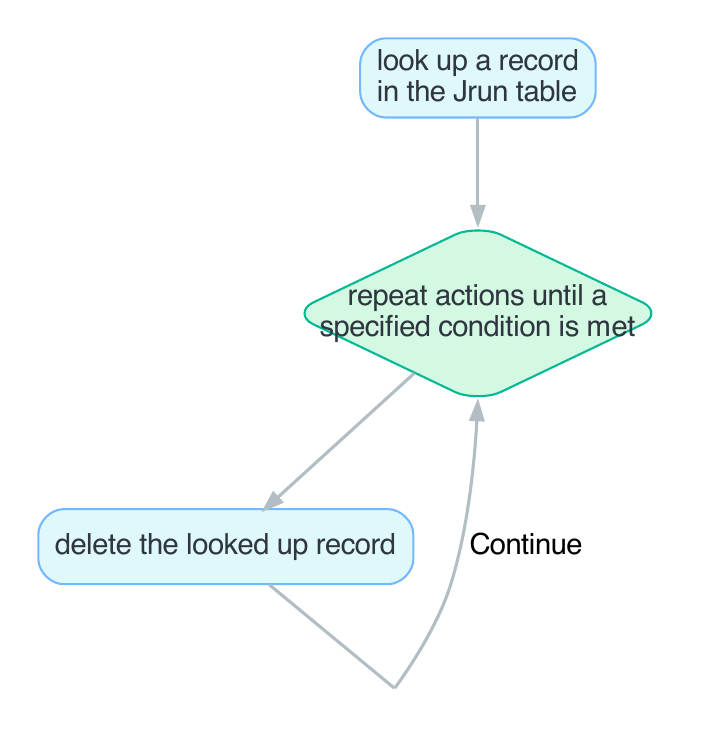}
        \caption{Qwen-2.5-VL-7B-Instruct (Finetuned)}
        \label{fig:sub4}
    \end{subfigure}

    \caption{\textbf{Error analysis across base and fine-tuned models on a \textsc{User Interface} workflow diagram.}
The Qwen-2.5-VL-Instruct model exhibits structural misinterpretations, confusing a \texttt{DOUNTIL} loop with an \texttt{IF} condition and incorrectly treating the \texttt{look\_up\_records} element as a trigger rather than a component. GPT-4o captures the correct flow logic but introduces an unnecessary empty trigger and an extra \texttt{end} component; it also selects an incorrect input table (not illustrated in the diagram). In contrast, the fine-tuned Qwen-2.5-VL model generates a fully accurate workflow, correctly handling both control logic and component definitions.}
    \label{fig:error_analysis_2}
\end{figure*}

\begin{figure*}[t]
    \centering
    
    \begin{subfigure}{\textwidth}
        \centering
        \includegraphics[width=\linewidth]{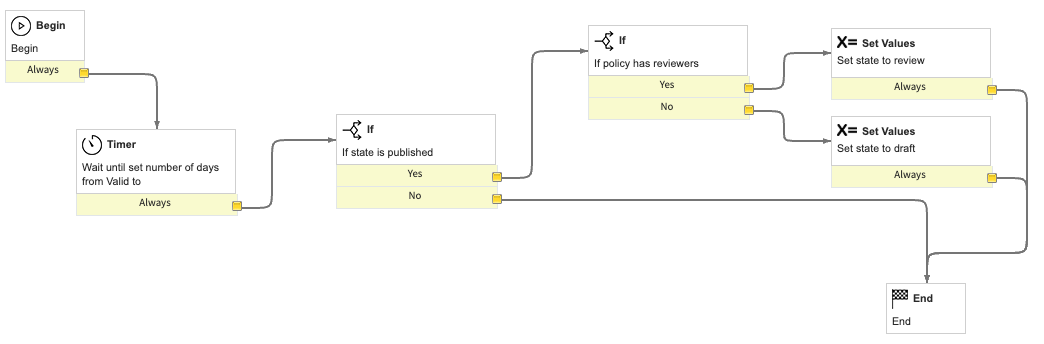}
        \caption{\textsc{Out-of-Distribution} Workflow Diagram}
        \label{fig:sub_top}
    \end{subfigure}

    \vspace{0.75em}

    \begin{subfigure}{0.32\textwidth}
        \centering
        \includegraphics[width=\linewidth]{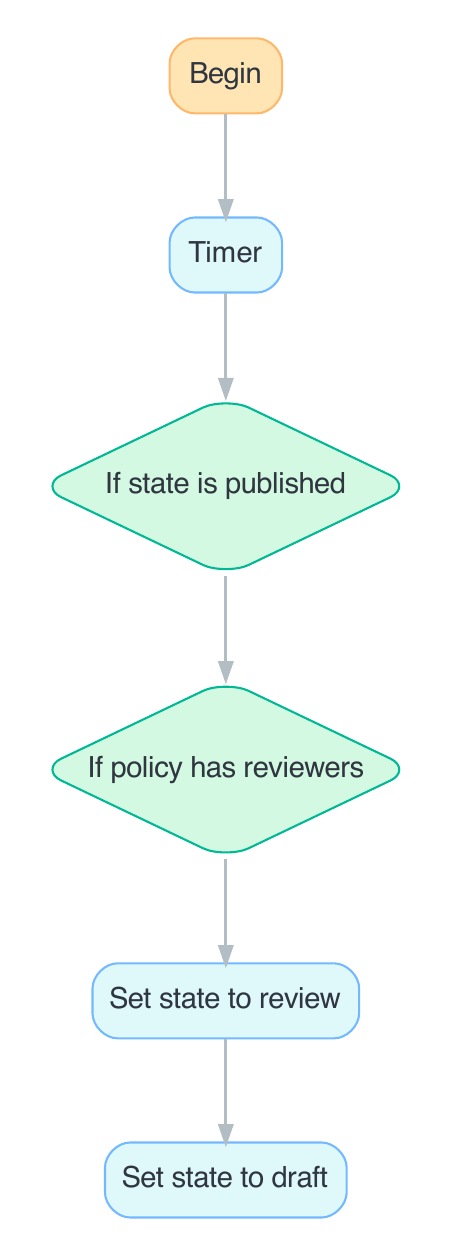}
        \caption{Pixtral-12B}
        \label{fig:sub_bottom1}
    \end{subfigure}
    \hfill
    \begin{subfigure}{0.32\textwidth}
        \centering
        \includegraphics[width=\linewidth]{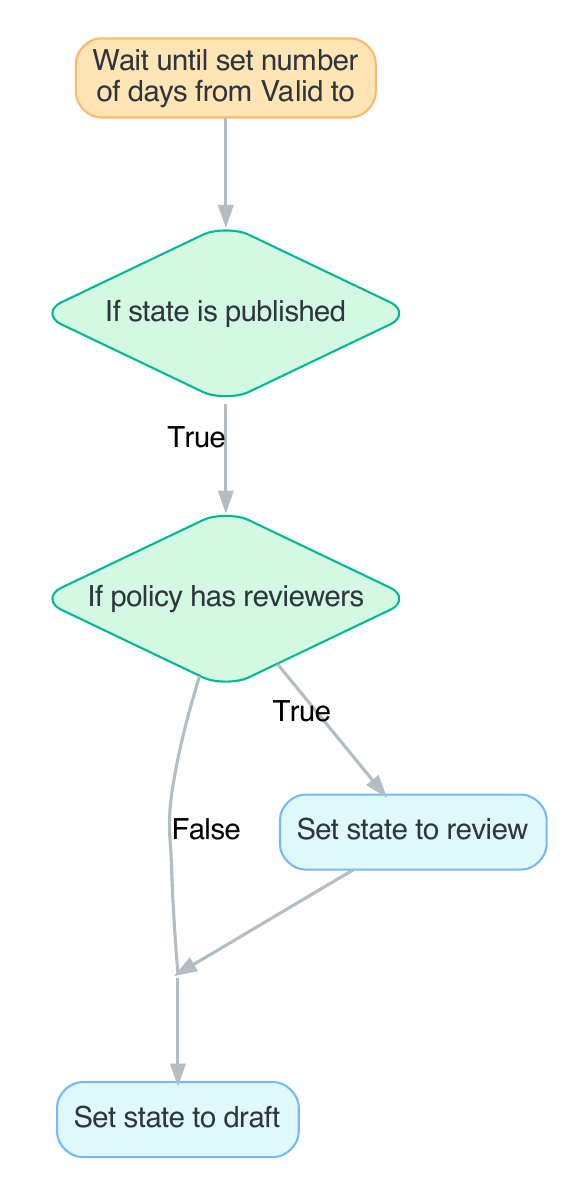}
        \caption{GPT-4o}
        \label{fig:sub_bottom2}
    \end{subfigure}
    \hfill
    \begin{subfigure}{0.32\textwidth}
        \centering
        \includegraphics[width=\linewidth]{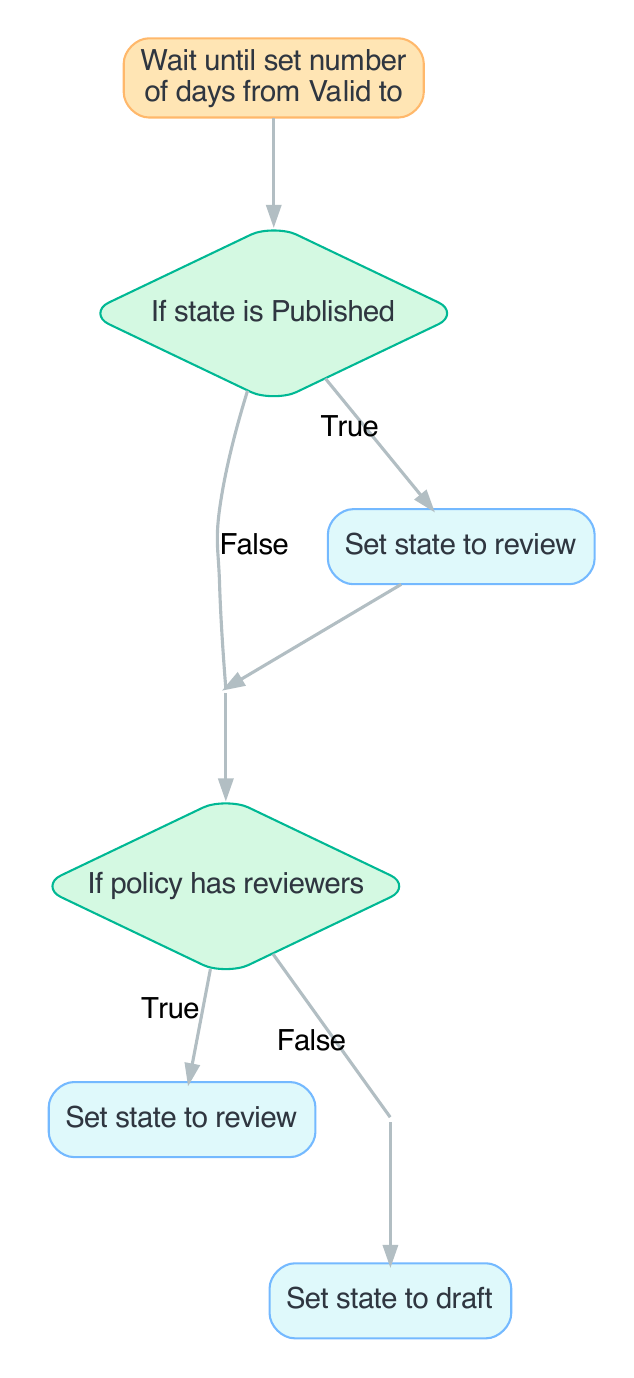}
        \caption{Pixtral-12B (Finetuned)}
        \label{fig:sub_bottom3}
    \end{subfigure}

    \caption{\textbf{Error analysis across base and fine-tuned models on an out-of-distribution workflow diagram.}
The Pixtral 12B model introduces a redundant \texttt{begin} trigger, incorrectly interprets the \texttt{IF} statements resulting in invalid control logic, and hallucinates \texttt{set\_values} components instead of generating the expected \texttt{update\_record} actions. GPT-4o produces a mostly correct flow but misclassifies the \texttt{TIMER} component as a trigger, and exhibits slightly incorrect logic in the second conditional, causing all records to be assigned the draft state while also hallucinating \texttt{set\_state} components instead of \texttt{update\_record} (not shown in diagram). The fine-tuned Pixtral 12B model correctly reflects condition-dependent record updates but still misidentifies \texttt{TIMER} component as a trigger and fails to terminate the flow when the state is not published.}
    \label{fig:error_analysis_3}
\end{figure*}

Figures \ref{fig:error_analysis_2} and \ref{fig:error_analysis_3} present more error failure modes of different models on various types of sample.

\newpage

\end{document}